\documentclass[10pt]{article}

\usepackage{geometry}  % set the margins to 1in on all sides
\usepackage[english]{babel}
\usepackage[utf8x]{inputenc}
\usepackage[T1]{fontenc}

\usepackage[dvipsnames]{xcolor}
\usepackage{paralist}
\usepackage{graphicx}
\usepackage{subcaption}
\usepackage{longtable} 
\usepackage{multirow}
\usepackage{listings}
\usepackage{makecell}
\usepackage{array}
\usepackage{float}
\usepackage{dsfont}
\usepackage{rotating}
\usepackage{booktabs}
\usepackage{enumerate}
\usepackage{tikz}
\usepackage{pgf}

\usepackage{amsmath}
\usepackage{amssymb}
\usepackage{amsthm}
\usepackage{bm}
\usepackage{mathtools}
\usepackage{mathrsfs}
\usepackage{algorithm}
\usepackage{algorithmic}

\usepackage{natbib}
\usepackage[
  colorlinks,
  citecolor=blue,
  linkcolor=red,
  anchorcolor=red,
  urlcolor=blue
]{hyperref}
\mathtoolsset{showonlyrefs}
\usepackage{authblk}
\usepackage{todonotes}
\usepackage{hyperref}
\usepackage{ying}
\long\def\comment#1{}

\let\hat\widehat
\let\tilde\widetilde

\usepackage{hyperref}
\def\##1\#{\begin{align}#1\end{align}}
\def\$#1\${\begin{align*}#1\end{align*}}

\makeatletter
\newcommand{\printfnsymbol}[1]{%
  \textsuperscript{\@fnsymbol{#1}}%
}
\makeatother

\usepackage{xcolor}

\title{Upper bounds on the Natarajan dimensions \\of some function classes}
\author{Ying Jin}
\affil{Department of Statistics, Stanford University}

\begin{document}

\maketitle

\begin{abstract}
The Natarajan dimension is a fundamental tool for characterizing multi-class PAC learnability, generalizing the Vapnik-Chervonenkis (VC) dimension from binary to multi-class classification problems. This work establishes upper bounds on Natarajan dimensions for certain function classes, including (i) multi-class decision tree and random forests, and (ii) multi-class neural networks with binary, linear and ReLU activations. These results may be relevant for describing the performance of certain multi-class learning algorithms. 
\end{abstract}

\section{Backgrounds}

Many tasks in statistical learning concern finding a good representation 
of the true relationship underlying the observations out of 
a perhaps huge family of functions.  
To this end, an intuitive and prominent approach is empirical risk minimization (ERM). 
Given i.i.d.~observations $\{(X_i,Y_i)\}_{i=1}^n\in\cX\times\cY$ 
and a loss function $\ell\colon \cY\times\cY\to \RR^+$, 
to learn a predictor $f\in\cF$ 
for $Y$ with the smallest loss,  
ERM selects the candidate with the smallest empirical prediction error: 
\$
\hat{f} = \argmin_{f\in \cF} \hat{L}_n(f),\quad 
\hat{L}_n(f) := \frac{1}{n} \sum_{i=1}^n \ell(f(X_i),Y_i), %\big(Y_i- f(X_i)\big)^2.
\$
% where $\ell(y',y)$ is a loss function describing the discrepancy 
% between a predicted value $y'$ to the true outcome $y$, such as 
% $\ell(y',y) = (y-y')^2$ for mean-squared-error in regression 
% or $\ell(y',y)=\ind\{y'=y\}$ for binary classification $Y\in\{0,1\}$, etc. 
Here, it is hoped that $\hat{L}_n(f)$ is a good estimate of the true prediction error 
$L(f)=\EE[\ell(f(X),Y)]$. 
The complexity of the function class $\cF$ 
is a crucial quantity that impacts the performance of such $\hat{f}$. 
The more complex $\cF$ is, 
the more likely it is that the smallest empirical risk occurs only by chance, 
and that the estimate for $\hat{f}$ is far from its true value. 
Understanding the complexity of function classes to characterize 
the learning performance is a fundamental task in statistical learning theory. 

\subsection{Natarajan dimension}
This paper studies the Natajaran dimension~\citep{natarajan1989learning}, 
a measure of the complexity in 
learning function classes
for \emph{multi-class classification} problems. 
It generalizes the well-known 
Vapnik-Chervonenkis (VC) dimension~\citep{vapnik2015uniform} 
for binary classification function classes. 
A closely related quantity, the graph dimension, is also defined below. 

\begin{definition}\label{def:nat}
Let $\cH$ be a class of functions $h\colon \cX\to \cY$, and let $S\subseteq \cX$. 
We say that $\cH$ G-shatters $S$ if there exists an $f\colon S\to \cY$ 
such that for every $T\subseteq S$, there exsits a $g\in \cH$ such that 
\$
\forall x\in T,~ g(x)=f(x),\quad \text{and}~~ \forall x\in S\backslash T, ~g(x)\neq f(x).
\$
We say that $\cH$ N-shatters $S$ if there exists $f_1,f_2\colon S\to \cY$ 
such that $f_1(x)\neq f_2(x)$ for all $x\in S$, 
and for every $T\subseteq S$, there exists some $g\in \cH$ such that 
\$
\forall x\in T,~ g(x)=f_1(x),\quad \text{and}~~ \forall x\in S\backslash T, ~g(x)=f_2(x).
\$
The graph dimension of $\cH$, denoted as $d_G(\cH)$, is the maximal cardinality of any set G-shattered by $\cH$. 
The Natarajan dimension of $\cH$, denoted as $d_N(\cH)$, is the maximal cardinality of 
any set N-shattered by $\cH$. 
\end{definition}

Both the graph dimension and the Natarajan dimension 
coincide with the VC dimension for $\cY=\{0,1\}$, and it 
is shown that 
$d_N(\cH)\leq d_G(\cH) \leq 4.67 \log_2(|\cY|) d_N(\cH)$~\citep{ben1992characterizations}. 
In this work, we only focus on the Natarajan dimension, 
and the results on the graph dimension can be easily obtained.

Similar to the VC dimension, 
the Natarajan dimension can be used to characterize 
the generalization of ERM -- more precisely, the true prediction error of 
the empirical risk minimizer -- using 
multi-class classification function classes~\citep{natarajan1989learning,ben1992characterizations,daniely2011multiclass}.     
However, unlike the extensively studied VC dimension, 
results on 
upper bounds on the Natarajan dimensions are relatively rare; 
existing results only cover linear function classes 
and reduction tree classes~\citep{daniely2011multiclass}. 
This work provides two more instances, 
decision trees (and random forests) and neural networks (fully-connected ones with linear, binary and ReLU activations). 
We study tree-based function classes in 
Section~\ref{sec:tree} and 
neural networks in Section~\ref{sec:nn}, 
while all proofs are in Section~\ref{sec:proof}. 

\subsection{Growth functions}

Our theoretical proof will follow the classical idea of bounding growth functions. 

\begin{definition}
Let $\cH$ be a class of functions $h\colon \cX\to \cY$. 
The growth function of $\cH$ is defined as 
\$
G(\cH, n) := \max_{x_1,\dots,x_n\in \cX} \Big| \big\{ \big(f(x_1),f(x_2),\dots,f(x_n)\big) \colon f \in \cH   \big\}    \Big|,
\$
which is the number of distinct realizations of $f\in \cH$ on any $n$ feature values, 
where for a finite set $A$, we write $|A|$ as the cardinality of $A$.
\end{definition}

More specifically, following the definition of Natarajan dimension, 
the functions in $\cH$ has at least $2^{d_{N}(\cH)}$ different configurations 
on $d_N(\cH)$ feature values in $\cX$. 
This fact allows us to utilize 
$2^{d_N(\cH)}\leq G(\cH, d_N(\cH))$ to 
obtain proper upper bounds on $d_N(\cH)$ for the function classes $\cH$ we 
study in this paper.

\subsection{Related work}
\label{subsec:related_work}

The Natarjan dimension and the graph dimension in Definition~\ref{def:nat} 
are both proposed by~\citep{natarajan1989learning} that generalize 
the Vapnik-Chervonenkis (VC) dimension~\citep{vapnik2015uniform} to 
multi-class problems. 
The Natarajan dimension is  
a crucial measure of PAC learnability; in particular,~\cite{daniely2011multiclass} 
shows that the sample complexity of 
PAC learning for multi-class classifications are bounded, 
in both directions, in terms of 
the Natarajan and the graph dimensions. 

This work is closely related to  the literature on 
establishing bounds on the Natarajan dimensions of popular function classes, 
including 
\cite{daniely2011multiclass} on generalized linear models and reduction trees, 
\cite{guermeur2010sample} on multi-class support vector machines, 
\cite{daniely2012multiclass} on one-versus-all, all-pairs, error-correcting-output-codes 
methods, etc. 
The current paper extends this line of work by establishing new bounds 
on popular tree-based and neural-network-based function classes. Notably, 
the decision trees we consider 
differ from the reduction trees in~\cite{daniely2011multiclass}, and 
should be viewed as  distinct function classes; see Remark~\ref{rmk:tree} for 
a detailed comparison.  
% the nodes of a reduction tree is a binary classifier from a specific class, 
% while our nodes take specific form of $x_j < c$ or $x_j\geq c$ 
% for some feature $j$; furthermore, 
% the leaf nodes of a reduction tree are one-to-one mapped to 
% all classes, while we allow several leaf nodes to map to a same class. 

The proof ideas in the work are inspired by several works 
in bounding the VC dimensions of binary-classification function classes. 
In particular, the technique in proving the results for neural network classifiers in 
Section~\ref{sec:nn} extends~\cite{sontag1998vc}.
The author was not aware of previous work on upper bounding the VC dimensions on decision trees with 
a given number of real-valued features and a given depth before a preliminary version 
of this work (as arXiv preprint 2209.07015). As discussed in an independent work~\citep{leboeuf2022generalization}, such results appear to be very recent.

\section{Tree-based function class}
\label{sec:tree}

Decision trees and random forests  
are popular tree-based machine learning methods 
that could be used for multi-class classification.   
This section provides upper bounds on the Natarajan dimensions 
of these classes. 

We first study the function class $\Pi_{L,d}^{\textrm{dtree}}$, 
each element of which is a 
depth-$L$ $d$-class decision tree.  
A depth-$L$ decision tree is a full binary tree, where 
each internal node $v$ is associated with 
a feature $i_v\in\{1,\dots,p\}$ and a threshold $\theta_v\in \mathbb{R}$, 
and each leaf node is associated with a class $k\in\{1,\dots, d\}$. 
For input $x\in \mathbb{R}^p$, the output is obtained by traversing a path of length $L-1$ from the root node to the leaf node. 
At each node $v$, 
if $x_{i_v}\leq \theta_{v}$ then we continue to its left child node, 
and to its right child node otherwise. 
The final classification is given by the class associated with the leaf node we arrive at. 

\begin{remark}\label{rmk:tree}
\cite{daniely2012multiclass} also studies the 
Natarajan dimension for decision trees, 
however, under a different definition: they
assume 
there is a bijection between the leaf nodes and the $d$ classes, 
and the internal nodes are from a general class of binary functions. 
Instead, we allow multiple leaf nodes to represent the same class, 
but consider more restricted binary classification rules for the internal nodes. 
We will obtain different upper bounds 
with different proof techniques. 
\end{remark}

The following theorem establishes upper bounds on the  Natarajan dimension of $\Pi_{L,d}^{\textrm{dtree}}$.

\begin{theorem}\label{lem:tree_forest_dim}
The Natarajan dimension of $\Pi_{L,d}^{\textrm{dtree}}$ 
with inputs from $\RR^p$ is no greater than $\mathcal{O}(L2^L\log(pd))$.  
\end{theorem}

We then consider the function class of random forests, 
denoted by $\Pi_{L,T,d}^{\textrm{forest}}$, 
each element of which is 
a random forest classifier $F(\cdot)$ consisting of $T$ depth-$L$ 
$d$-class decision trees $f_j(\cdot)$, $j=1,\dots,T$. 
Given any $x\in \RR^p$, 
the output of a random forest is given by $F(x) = \mathop{\textrm{argmax}}_{1\leq k\leq d} \sum_{j=1}^T \mathbf{1}\{f_j(x)=k\}$, the most-frequently predicted class among all $T$ trees. 
Its Natarajan dimension can be bounded  
as follows.  

\begin{theorem}\label{lem:forest_dim}
The Natarajan dimension  
of $\Pi_{L,T,d}^{\textrm{forest}}$ 
with inputs from $\RR^p$ is no greater than $\mathcal{O}(LT2^L\log(pd))$. 
\end{theorem}

As we discussed in Section~\ref{subsec:related_work}, 
upper bounds on the VC dimension of 
decision trees with real-valued features appear to be very recent results~\citep{leboeuf2022generalization}. 
In particular, the independent recent work of~\cite{leboeuf2022generalization} shows that 
the VC dimension of  
binary decision trees with $L_T$ leaves for $p$ real-valued 
features and the same splitting rules 
as ours is $\cO(L_T\log(L_t p))$. 
Since the number of leaves is $2^L-1$ for a tree of depth $L$, 
our bound $\cO(L 2^L\log(pd))$ in Theorem~\ref{lem:tree_forest_dim} 
only pays a price of $\log(d)$ 
for $d$-class classification compared with the results in~\cite{leboeuf2022generalization}, 
while the rate in the number of leaves is the same.

\section{Neural network function class}
\label{sec:nn}

For a number $d$ of actions, a multiple classification neural network has $d$ outputs in the final layer and constructs a classification by taking the maximum over these outputs. 

\subsection{Neural network function class with binary and linear activations}

We first consider  $\Pi_{p,S}^{\textrm{binary}}$, 
a neural network function class with a fixed structure $S$ of $p$ parameters, 
where  
all activation functions are either binary or linear, 
which generalizes the setting in~\citep{sontag1998vc}.

The fixed structure $S$ consists of $L$ layers, 
where the $\ell$-th layer has $n_{\ell}$ nodes, $\ell\in \{1,\dots,L\}$. 
We denote the $j$-th node in layer $\ell$ as $\texttt{Node}_{\ell,j}$, 
and denote $\mathcal{N}_{\ell,j}$ as the set of nodes in layer $\ell-1$ 
that are  connected to $\texttt{Node}_{\ell,j}$, whose size is 
$m_{\ell,j} = |\mathcal{N}_{\ell,j}|$.  
There is one real-valued paremeter for each pair of connected nodes 
in adjacent layers. 
We define the set of parameters of the network 
as $w = \{w_{\ell,j,s}\}_{1\leq s\leq m_{j,\ell}, 1\leq \ell<L}$ (excluding those weights for the last output layer), which 
we assume is of a size smaller than $p$. 

Each element in $ \Pi_{p,S}^{\textrm{binary}}$ is a  
feed-forward neural network; 
given any input $x\in \RR^m$, it 
 outputs 
$f(x;\,w)$  as follows.    
The input layer takes $x\in \mathbb{R}^m$ from $m$ input nodes, each for one feature. 
In each hidden layer $\ell\in\{1,\dots,L-1\}$, $\texttt{Node}_{\ell,j}$ 
performs a linear combination to the
outputs from each node in $\mathcal{N}_{\ell,j}$ in the previous layer, 
using $m_j$ parameters $w_{\ell,j,1},\dots,w_{j,m_j}\in \mathbb{R}$.
To be specific, the output of $\texttt{Node}_{\ell,j}$ is 
$$
f^{(\ell)}_j(x) = \sigma\bigg( \sum_{s\in \mathcal{N}_{\ell,j}} w_{\ell,j,s} \cdot f^{(\ell-1)}_s(x) \bigg),
$$
where $\sigma(\cdot)$ is either the binary activation $\sigma(z)=\mathbf{1}\{z>0\}$ or the linear activation $\sigma(z) = z$ for $z\in \mathbb{R}$. The last layer has $d$ nodes, and is fully-connected to the 
last hidden layer with an identity activation. The output of the neural network is 
$$
f(x;\,w) = \mathop{\textrm{argmax}}_{1\leq k\leq d}~ \Bigg\{\sum_{s=1}^{n_{L-1}} w_{L,k,s} \cdot f^{(L-1)}_{s}(x)\Bigg\}.
$$  

The following theorem provides upper bounds on 
Natarajan dimensions of such function classes. 

\begin{theorem}\label{lem:nn_binary_dim}
The Natarajan dimension of $\Pi_{p,S}^{\textrm{binary}}$ described 
above  
is upper bounded by $\mathcal{O}(d\cdot p^2)$. 
\end{theorem}
 
Several seminal early works have established 
the VC dimension of neural networks. 
The textbook result in~\cite{shalev2014understanding} provides 
an upper bound of $\cO(p\log p)$ for neural networks with $p$ parameters 
and all binary activation functions 
$\sigma(z)=\mathbf{1}\{z>0\}$. 
\cite{sontag1998vc} provides an upper bound of $\cO(p^2)$ 
for neural networks with $p$ parameters and binary or linear activation functions 
as considered in this subsection. 
By comparing this two results, we see that the allowing for linear activation functions 
adds a factor of $p$ to the VC dimension (from the rate of $p\log p$ to $p^2$). 
Also, moving from binary to $d$-class classification 
adds a multiplicative factor of $d$ to the upper bound when 
comparing Theorem~\ref{lem:nn_binary_dim} with~\cite{sontag1998vc}. 
This is because the $d$-output neural network structure leads to 
a power-$d$ factor in the growth function using our current proof technique. 
It will be interesting to see whether such dependence on $d$ can be further 
sharpened. 

\subsection{Neural network function class with ReLU activations} 

We now consider $\Pi_{p,S}^{\textrm{ReLU}}$, 
the class of multiple classification neural networks with a given structure $S$, 
which contains at most $p$ parameters in intermediate layers and $d$ final outputs, and the activation functions are either binary, linear or ReLU, i.e., $\sigma(z)=z$ or $\sigma(z)=\mathbf{1}\{z>0\}$ or $\sigma(z)=z\mathbf{1}\{z>0\}$. 
The definition of structure $S$ is the same as in the 
preceding subsection; the only difference is that 
the activation functions for internal nodes can now be more general. 

The Natarajan dimension of this function class 
is upper bounded as in the following theorem. 

\begin{theorem}\label{lem:nn_relu_dim}
The Natarajan dimension of $\Pi_{p,S}^{\textrm{ReLU}}$  
described above 
is upper bounded by $\mathcal{O}(d\cdot p^2)$. 
\end{theorem}

Our bound in Theorem~\ref{lem:nn_relu_dim} is of the same order 
as that in Theorem~\ref{lem:nn_relu_dim}. Intuitively, this is because 
the ReLU activation can be viewed as a combination of binary and linear activation, 
which does not significantly increase the growth function compared with 
networks with the latter two activation functions. 

Our theoretical analysis for the Natarajan dimensions of neural network function 
class is largely inspired by the framework of~\cite{sontag1998vc}, which 
expresses neural network outputs as depending on linear combinations 
of binary values and original features, where the linear cofficients 
are further polynomials of the parameters of a bounded degree. 

\begin{remark}
The price we pay for the number of classes $d$ in 
neural networks is a multiplicative factor of $d$, 
which is higher than the $\log(d)$ factor for decision trees 
and random forests. 
Although we only provide an upper-bound analysis, 
such comparison potentially indicates that 
the complexity of multi-class classification with 
neural networks might be higher than that of tree-based function classes. 
\end{remark}

\section{Technical proofs}
\label{sec:proof}

In this section, we provide the proofs for all results in this work.

\subsection{Proof of Theorem~\ref{lem:tree_forest_dim} } \label{app:proof_dtree}

Given a set of inputs $\{x_1,\dots,x_n\}\in \mathbb{R}^p$, we first bound the number of configurations of the output on $\{x_1,\dots,x_n\}$ by a decision tree of depth $L$. Denote the growth function on $x_1,\dots,x_n$ as 
$$
g(\Pi_{L,d}^{\textrm{dtree}},n\given x_1,\dots,x_n) = \Big| \big\{ \big(f(x_1),f(x_2),\dots,f(x_n)\big) \colon f \in \Pi_{L,d}^{\textrm{dtree}}   \big\}    \Big|.
$$
We sort and denote internal nodes as $v\in \{1,\dots,V\}$, where $V=2^{L-1}-1$ is the total number of internal nodes. 
We denote the corresponding feature as ${i_v}$ and the threshold as $\theta_{v}$, 
which vary with the tree $f\in\Pi_{L,d}^{\textrm{dtree}}$. 
We index the set of leaf nodes with $\{1,\dots,2^{L-1}\}$, and 
 each leaf node $1\leq l \leq 2^{L-1}$ 
represents a class $k_l\in\{1,\dots,d\}$.  
In this way, given any input $x_i\in \mathbb{R}$, 
the output  from a tree $f\in \Pi_{L,d}^{\textrm{dtree}}$ is 
determined by the vector of queries $(\mathbf{1}\{x_{i,i_v}\leq \theta_n\})_{v=1}^V \in \mathbb{R}^V$ at all the $V$ internal nodes, 
as well as $\{k_l\colon 1\leq l \leq 2^{L-1}\}$, 
the assignment of classes to all leaf nodes. 
To better represent the classification model, 
for a decision tree $f\in \Pi_{L,d}^{\textrm{dtree}}$, we let $I_{f,v}(x)  =\mathbf{1}\{x_{i,i_v}\leq \theta_v\}$ be the query function at node $v$,  
$I_f(x) = (\mathbf{1}\{x_{i,i_v}\leq \theta_n\})_{v=1}^V $ be the vector of queries at internal nodes, and $L(f) = (k_l)_{l=1}^{2^{L-1}}$ be the vector of leaf node assignments.

Since $f(x)$ is fully decided by $I_f(x)$ and $L(f)$, we obtain an upper bound on the growth function that 
\begin{align*}
g(\Pi_{L,d}^{\textrm{dtree}},n\given x_1,\dots,x_n) &\leq \Big| \big\{ \big(I_f(x_1),I_f(x_2),\dots,I_f(x_n), L(f)\big)\in \{0,1\}^{n\times V}\times \{1,\dots,d\}^{2^{L-1}} \colon f \in \Pi_{L,d}^{\textrm{dtree}}   \big\}    \Big| \\
&\leq  \Big| \big\{ \big(I_f(x_1),I_f(x_2),\dots,I_f(x_n)\big)\in \{0,1\}^{n\times V}\colon f \in \Pi_{L,d}^{\textrm{dtree}}   \big\}    \Big| \\
&\qquad \times \Big| \big\{  L(f) \in   \{1,\dots,d\}^{2^{L-1}} \colon f \in \Pi_{L,d}^{\textrm{dtree}}   \big\}    \Big|.
\end{align*}
The second term is upper bounded by the cardinality of the image as
$$
\Big| \big\{  L(f) \in   \{1,\dots,d\}^{2^{L-1}} \colon f \in \Pi_{L,d}^{\textrm{dtree}}   \big\}    \Big| \leq d^{2^{L-1}}.
$$
For the first term, the total number of different configurations of $I_f$ is upper bounded by the product of those of all $I_{f,v}$, namely,
\begin{align*}
&\Big| \big\{ \big(I_f(x_1),I_f(x_2),\dots,I_f(x_n)\big)\in \{0,1\}^{n\times V}\colon f \in \Pi_{L,d}^{\textrm{dtree}}   \big\}    \Big| \\
&\leq \prod_{v=1}^V \Big| \big\{ \big(I_{f,v}(x_1),I_{f,v}(x_2),\dots,I_{f,v}(x_n)\big)\in \{0,1\}^{n}\colon f \in \Pi_{L,d}^{\textrm{dtree}}   \big\}    \Big|\\
&= \Big| \big\{ \big(I(x_1),I (x_2),\dots,I (x_n)\big)\in \{0,1\}^{n}\colon I \in \mathcal{I}   \big\}    \Big|^V,
\end{align*}
since each query function $I_{f,v}$ belongs to the function class 
$$
\mathcal{I} = \Big\{ I\colon \mathbb{R}^p \to \{0,1\}\colon  I(x) = \mathbf{1}\{x_i\leq \theta\},~i\in \{1,\dots,p\},~\theta\in \mathbb{R} \Big\}.
$$
Moreover, for any function $I\in \mathcal{I}$ of the form 
$I(x)=\mathbf{1}\{x_s\leq \theta\}$ for some feature $s\in\{1,\dots,p\}$, 
there are at most $n+1$ different classifications on the $n$ samples: 
one could sort all possible classification results 
into a sequence of size $n+1$ (perhaps with recurring members), 
such that 
the $\ell$-th of result classifies the $\ell$ sample with smallest $x_s$ as positive and others as negative. Hence
$$
\Big| \big\{ \big(I(x_1),I (x_2),\dots,I (x_n)\big)\in \{0,1\}^{n}\colon I \in \mathcal{I}   \big\}    \Big| \leq p(n+1).
$$
Putting this together, we have $g(\Pi_{L,d}^{\textrm{dtree}},n\given x_1,\dots,x_n) \leq (p(n+1))^{2^{L-1}-1} \cdot d^{2^{L-1}}$. 

Now suppose $\{x_1,\dots,x_N\}$ are N-shattered by $\Pi_{L,d}^{\textrm{dtree}}$. By definition, policies in $\Pi_{L,d}^{\textrm{dtree}}$ have at least $2^N$ different configurations on $\{x_1,\dots,x_N\}$, leading to 
$$
2^N \leq (p(N+1))^{2^{L-1}-1} \cdot d^{2^{L-1}}.
$$
Taking logarithm of both sides yields
$
N\log 2 \leq 2^L \log(pd) + 2^L \log N
$, which further gives $N =  \mathcal{O}(L2^L\log(pd))$. This proves the upper bound of Natarajan dimension of $\Pi_{L,d}^{\textrm{dtree}}$. \hfill $\square$

\subsection{Proof of Theorem~\ref{lem:forest_dim}}

Suppose $\{x_1,\dots,x_N\}$ is N-shattered by $\Pi_{L,T,d}^{\textrm{forest}}$. 
Since the output of a policy $f\in \Pi_{L,T,d}^{\textrm{forest}}$ is fully decided by the outputs of $T$ decision trees $f_1,\dots,f_T\in \Pi_{T,d}^{\textrm{dtree}}$, the number of different configurations on $\{x_1,\dots,x_N\}$ by policies in $\Pi_{L,T,d}^{\textrm{forest}}$ is upper bounded by $g(\Pi_{L,d}^{\textrm{dtree}},N\given x_1,\dots,x_N)^T$. Hence, 
$$
2^N \leq g(\Pi_{L,d}^{\textrm{dtree}},N)^T \leq (p(N+1))^{T(2^{L-1}-1)} \cdot d^{T\cdot 2^{L-1}}.
$$
This implies $N \leq  \mathcal{O}(LT2^L\log(pdT))$, which completes the proof of Theorem~\ref{lem:tree_forest_dim}. \hfill $\square$

\subsection{Proof of Theorem~\ref{lem:nn_binary_dim}} \label{app:proof_nn_binary}

% Our proof is a generalization of Proposition 1 in \cite{sontag1998vc}.  
Firstly, we fix any $f\in \Pi_{p,S}^{\textrm{binary}}$. Note that 
the final output $f(x)$  
is fully determined by the $d(d-1)/2$ binary functions 
$\{b_{k,k'}(\cdot)\colon 1\leq k < k; \leq d\}$, where 
$$
b_{k,k'}(x) = \mathbf{1}\Bigg\{ \sum_{s=1}^{n_{L-1}} \theta_{L,k,s} \cdot f^{(L-1)}_{s}(x) - \sum_{s=1}^{n_{L-1}} \theta_{L,k',s} \cdot f^{(L-1)}_{s}(x) >0 \Bigg\}.
$$
For simplicity, we assume there are no ties among the $k$ outputs; otherwise 
one could break the ties without changing our results.

We now proceed to analyze the outputs $f_j^{(\ell)}$ in the hidden layers 
for $\ell\leq L-1$. Similar to the idea in~\cite{sontag1998vc}, 
we will show that they can all be expressed 
as linear combinations of binary values 
and $x_1,\dots,x_m$.
To begin with, we note that the activation functions $\sigma(\cdot)$ are either binary $\sigma(z)=\ind\{z>0\}$ or linear $\sigma(z)=z$, hence the 
outputs at layer $\ell=1$ are either binary-valued, or some linear function of $x$. 
That is, we either have $f_j^{(1)}(x)\in\{0,1\}$, or 
$
f_j^{(1)}(x) = x^\top \alpha_j^{(1)}
$
for some $\alpha_j^{(1)}\in \RR^m$. 
We let $f_s^{(\ell)}(x)=\sigma(g_s^{(\ell)}(x))$, i.e., 
we denote $g_s^{(\ell)}(x)$ as the intermediate output 
before applying the activation function for node $s$ at layer $j$. 
By definition, $g_s^{(2)}(x)$ is some linear combination of 
$\{f_j^{(1)}(x)\}$, the node outputs at layer $1$. Therefore, 
for any $s$, $g_s^{(2)}(x)$ is a linear combination 
of some binary values and some linear function of $x$. 
Letting $I(z)=\mathbf{1}\{z>0\}$. We can write 
\$
g_s^{(2)}(x) &= \sum_{j\in \cN_{2,s},\sigma(z)=\mathbf{1}\{z>0\}} w_{2,s,j} I(x^\top \alpha_j^{(1)}) 
+ \sum_{j\in \cN_{2,s},\sigma(z)=z} w_{2,s,j} x^\top \alpha_j^{(1)} \\ 
&= \sum_{j\in \cN_{2,s},\sigma(z)=\ind\{z>0\}} w_{2,s,j} I( \theta_{1,j,1}x_1 + 
\theta_{1,j,1}x_2+\cdots +\theta_{1,j,m}x_m ) \\ 
&\qquad \qquad + w_{2,s,j}\big(  \theta_{1,j,1}x_1 + 
\theta_{1,j,2}x_2+\cdots +\theta_{1,j,m}x_m \big) \\ 
&:= \sum_{j\in \cN_{2,s},\sigma(z)=\ind\{z>0\}} \theta_{2,s,j} I( \theta_{1,j,1}x_1 + 
\theta_{1,j,1}x_2+\cdots +\theta_{1,j,m}x_m )  \\ 
&\qquad \qquad + \theta_{2,s,n_{2,j}-m+1} x_1 + \cdots \theta_{2,s,n_{2,s}} x_m
\$
for some $\theta_{1,j,1},\dots,\theta_{1,j,m}$ 
and  $\theta_{2,s,1},\dots,\theta_{2,s,n_{2,s}}\in \RR$
that can be derived from the network structure and $\{w_{j,\ell,s}\}$, 
where $n_{2,j}$ are the total number of $\theta_{2,s,\cdot}$ needed in such expression. 
Following this rule, we can write 
\$
g_s^{(3)}(x) &= w_{3,s,1} I\big(\underbrace{\theta_{2,1,1}I(\cdots) + \theta_{2,1,2} I(\cdots) +\cdots 
+ \theta_{2,1,n_{2,1}-m+1} x_1 + \cdots + \theta_{2,1,n_{2,1}}x_m}_{g_{1}^{(2)}(x) \textrm{ supposing the activation function at this node is }\sigma(z)=\mathbf{1}\{z>0\}}\big)  \\ 
&\qquad + \underbrace{w_{3,s,2} I\big(\cdots)}_{\textrm{other }g_j^{(2)} \textrm{ with activation function }\sigma(z)=\mathbf{1}\{z>0\}}  + \cdots + \underbrace{\cdots}_{\textrm{other }g_j^{(2)} \textrm{ with activation function }\sigma(z)=z}  \\ 
&\qquad + w_{3,s,m_{3,s}} \big(  \underbrace{\theta_{2,1,1}I(\cdots) + \theta_{2,1,2} I(\cdots) +\cdots 
+ \theta_{2,1,n_{2,1}-m+1} x_1 + \cdots + \theta_{2,1,n_{2,1}}x_m }_{g_{j}^{(2)}(x) \textrm{ supposing the activation function at this node is }\sigma(z)=z} \big) \\ 
&= \theta_{3,s,1} I(\cdots) + \theta_{3,s,2} I(\cdots) + \cdots 
+ \theta_{3,s,n_{3,s}-m+1} x_1 + \cdots + \theta_{3,s,n_{3,s}} x_m
\$
for some $\theta_{3,s,1},\dots, \theta_{3,s,n_{3,s}}\in \RR$ 
and $n_{3,s}$ is the number of such coefficients. 
In the last expression, each $I(\cdots)$ represents a node with binary activation, 
and inside the argument is another linear combination of binary values 
and $x_1,\dots,x_m$. 
So on and so forth, we have 
%  layer $L-1$ that are 
% shared by all $k$ outputs. 
% We let $f_s^{(L-1)}(x)=\sigma(g_s^{(L-1)}(x))$, i.e., 
% we denote $g_s^{(L-1)}(x)$ as the output before applying the activation function. 
% Note that the outputs $g_s^{(L-1)}(x)$ do not vary with $k$.  
% 
% Since the activation functions $\sigma(\cdot)$ are either binary $\sigma(z)=\ind\{z>0\}$ or linear $\sigma(z)=z$, 
% we know that $g_s^{(L-1)}(x)$ is a combination of some binary 
% outputs and some linear outputs from layer $L-2$. 
% 
\begin{align*}
g^{(L-1)}_{s}(x) &=  \theta_{L-1,s,1}I\big(\theta_{L-2,1,1}I(\cdots) + \theta_{L-2,1,2}I(\cdots) + \cdots \\
&\qquad \qquad \qquad + \theta_{L-2,1,n_{L-1,1}-m+1} x_1 + \cdots + \theta_{L-2,1,n_{L-1,1}} x_m \big) + \theta_{L-1,s,2} I(\cdots) + \cdots \\
&\qquad + \theta_{L-1,s,n_{L,s}-m+1} x_1 + \cdots + \theta_{L-1,s,n_{L,s}} x_m
\end{align*}
for some $\theta_{L-1,s,1},\dots,\theta_{L-1,s,n_{L-1,s}}\in \RR$. 
% where $I(z)=\mathbf{1}\{z>0\}$. Here $n_{L-1,s,1}$ is the number of terms that are included in the summation, including binary terms and linear combination of $x_1,\dots,x_n$. 
Put another way, the outputs $g_s^{(L-1)}(x)$ can be written as the linear combination of outputs of some binary functions and original features, and the arguments to these binary functions are again some linear combination of binary functions (computed from preceding layers) and the original features, and so on. Also, the coefficients $\theta_{\ell,j,s}$ are all polynomials of $x_1,\dots,x_m$ and $w_1,\dots,w_p$, whose degrees are no larger than $p$ since there are at most $p$ layers. Similarly, the function $b_{k,k'}(x)$ can be written as a binary function $b_{k,k'}(x)=I(c_{k,k'}(x))$, where $c_{k,k'}(x)$ is a linear combination of several binary functions decided by previous layers and the original features, for which the linear coefficients are polynomials of $x_1,\dots,x_m$, the original parameters $w_1,\dots,w_p$ and $\{w_{L,k,s}\}_{1\leq k\leq d,1\leq s\leq n_{L-1}}$ of degree $\leq p+1$.

Assume there are $p_1$ binary nodes in intermediate layers. 
Then those binary functions have at most $2^{p_1}$ configurations. 
Therefore, for any input $x,x'\in \mathbb{R}^m$ and parameters $w,w'\in \mathbb{R}^p$, we will have $f(x;\,w)=f(x';\,w')$ if the set of  binary functions 
\begin{align*}
\mathcal{I} := &\Big\{ I(\theta_{\ell,j,1} b_1 + \cdots \theta_{\ell,j,p_1} b_{p_1} + \theta_{\ell,j,n_{\ell,j}-m+1} x_1 + \cdots + \theta_{\ell,j,n_{\ell,j}} x_m  ) \colon \\
&\quad \qquad (b_1,\dots,b_{p_1})\in \{0,1\}^{p_1}, 1\leq\ell\leq L, 1\leq j\leq n_\ell  \Big\}
\end{align*}
take the same value, where $\theta_{\ell,j,s}$ are fixed polynomials of the parameters $w_1,\dots,w_p$, features $x_1,\dots,x_m$ and $\{w_{L,k,s}\}_{1\leq k\leq d,1\leq s\leq n_{L-1}}$ of degree $\leq p+1$. Note that there are at most $(d^2 +p) 2^{p}$ functions in $\mathcal{I}$, because for each configuration of $(b_1,\dots,b_{p_1})$, there are at most $p$ binary nodes in the intermediate layers, and no more than $d^2$ comparisons 
among the $d$ classes in the last layer. In other words,
\begin{align*}
\mathcal{I}   = \Big\{ I\big(P_r (x, w_1,\dots,w_p, \{w_{L,k,s}\}_{1\leq k\leq d,1\leq s\leq n_{L-1}} )\big)\colon r=1,\dots, R  \Big\},
\end{align*}
where $R\leq (p+d^2)2^p$ and each $P_r(\cdot)$ is a polynomial with degree $\leq p+1$. If we view the input $x$ as fixed, the above set of functions is 
\begin{align*}
\mathcal{I}(x) := \Big\{ I\big(P_r (x, w_1,\dots,w_p, \{w_{L,k,s}\}_{1\leq k\leq d,1\leq s\leq n_{L-1}} )\big)\colon r=1,\dots, R  \Big\},
\end{align*}
and each $P_r(\cdot)$ is a polynomial of degree $\leq p+1$ 
of no greater than $p(1+d)$ variables 
\$\big\{w_1,\dots,w_p, \{w_{L,k,s}\}_{1\leq k\leq d,1\leq s\leq n_{L-1}}\big\},\$
since the parameters for the last layer is at most $p\cdot d$. 

Let $x_1,\dots,x_N$ be N-shattered by the function class $\Pi_{p,S}^{\textrm{binary}}$, and let $w^{(1)},\dots, w^{(M)}$, $M=2^N$ be the parameters that witness the shattering. Consider the classifications
\begin{equation*}
\begin{pmatrix}
f(x_1;\, w^{(1)}), & ~f(x_2;\, w^{(1)}), & ~\cdots, & ~f(x_N;\, w^{(1)}) \\
f(x_1;\, w^{(2)}), & ~f(x_2;\, w^{(2)}), & ~\cdots, & ~f(x_N;\, w^{(2)}) \\
\vdots & \vdots & &  \vdots \\
f(x_1;\, w^{(M)}), & ~f(x_2;\, w^{(M)}), & ~\cdots, & ~f(x_N;\, w^{(M)})
\end{pmatrix}.
\end{equation*}
By the definition of N-shattering, every two rows in the above matrix are distinct. Thus for each $j\neq j'$, there exists some $i\in \{1,\dots,N\}$ such that $f(x_i;\, w^{(j)})\neq f(x_i;\, w^{(j')})$. By above arguments, there exists some binary function in $\mathcal{I}$ that takes different values on $(x_i,w^{(j)})$ and $(x_i, w^{(j')})$. 
Then there must exist some $1\leq r\leq R$ such that the signs of $P_r(x_i, w^{(j)})$ and $P_r(x_i,w^{(j')})$ are different. In other words, each $j\in \{1,\dots,M\}$ gives a unique configuration of the signs of the $N\cdot R$ polynomials in $\mathcal{I}(x_1),\dots, \mathcal{I}(x_N)$.  The following lemma establishes an upper bound for 
the number of such configurations, which is a re-statement of Corollary 2.1 in \cite{goldberg1995bounding}.

\begin{lemma}\label{lem:nn_config}
Let $\{P_1,\dots,P_{\tilde{R}}\}$ be $\tilde{R}$ polynomials of degree at most $\tilde{p}$ in $\tilde{n}$ real variables with $\tilde R\geq \tilde n$, then the number of different configurations of signs to the $\{P_1,\dots,P_{\tilde R}\}$ is at most $(8e\tilde{p}\tilde{R}/\tilde{n})^{\tilde{n}}$. 
\end{lemma}

Utilizing Lemma~\ref{lem:nn_config} with $\tilde{R}=N R$, $\tilde{p}=p+1$ and $\tilde{n}=p(1+d)$, the number of different configurations on the $N\cdot R$ polynomials is upper bounded as
$$
M = 2^N \leq \Big(\frac{8e(p+1)\cdot N(p+d^2)\cdot 2^p}{p(1+d)}\Big)^{p(1+d)}.
$$
Taking logarithm we have 
$$
N \leq p(1+k) \log\big( 8e\cdot (p+1) N d\cdot  2^p  \big),
$$
hence $N\leq  \mathcal{O} (dp^2)$, 
which completes the proof of Theorem~\ref{lem:nn_binary_dim}. 
\hfill $\square$

\subsection{Proof of Theorem~\ref{lem:nn_relu_dim}} \label{app:proof_nn_relu}
The proof of Theorem~\ref{lem:nn_relu_dim} is similar to that of Theorem~\ref{lem:nn_binary_dim}, except that we consider the configuration on signs of a slightly different set of polynomial functions. The formal definition of the structure is the same as the previous case, except that some of the activation functions are $\sigma(z)=z\mathbf{1}\{z>0\}$. We use the same notations for nodes and weights as in the previous case.  

For $\texttt{Node}_{j,\ell}$, we suppose its output is $f_j^{(\ell)}(x) = \sigma\big(g_{j}^{(\ell)}(x)\big)$, where 
$g_{j}^{(\ell)}(x)$ is the quantity before 
applying the activation function, i.e., the linear combination of the outputs of nodes in the preceding layer. 
If $\texttt{Node}_{j,\ell}$ has binary activation, then it appears in the formula for nodes in latter layers as $0$ or $1$ (see arguments in the proof of Theorem~\ref{lem:nn_binary_dim}). If $\texttt{Node}_{j,\ell}$ has ReLU activation, then it appears as either $0$ or $g_j^{(\ell)}(x)$, the linear combination itself. Recall that $I(z)=\mathbf{1}\{z>0\}$.  For any inputs $x,x'\in \mathbb{R}^m$ and parameters $w,w'\in \mathbb{R}^{p(1+d)}$ including parameters in the last layer, we have $f(x;\, w)=f(x';\, w')$ if the set of binary functions 
\begin{align*}
\mathcal{I} = &\Big\{ I(\theta_{\ell,j,1} b_1 + \cdots \theta_{\ell,j,p_1} b_{p_1} + \theta_{\ell,j,n_{\ell,j}-m+1} x_1 + \cdots + \theta_{\ell,j,n_{\ell,j}} x_m  ) \colon \\
&\quad \qquad (b_1,\dots,b_{p_1})\in \{0,1\}^{p_1}, 1\leq\ell\leq L, 1\leq j\leq n_\ell  \Big\}
\end{align*}
all have the same sign on $(x, w)$ and $(x',w')$. Here $p_1$ is the total number of binary and ReLU nodes, so that $p_1\leq p$. 
Also, each $\theta_{\ell,j,s}$ is a polynomial in all entries of $x,w$ of degree $\leq p+1$. Viewing them as functions of parameters, the set of functions is 
\begin{align*}
  \mathcal{I}(x) := \Big\{ I\Big(P_r\big(x, w_1,\dots,w_p, \{w_{L,k,s}\}_{1\leq k\leq d,1\leq s\leq n_{L-1}}\big)\Big)\colon r=1,\dots, R  \Big\},
\end{align*}

Let $x_1,\dots,x_N$ be N-shattered by the function class $\Pi_{p}^{\textrm{ReLU}}$, and let $w^{(1)},\dots, w^{(M)}$, $M=2^N$ be the parameters that witness the shattering. Consider the classifications
\begin{equation*}
\begin{pmatrix}
f(x_1;\, w^{(1)}), & ~f(x_2;\, w^{(1)}), & ~\cdots, & ~f(x_N;\, w^{(1)}) \\
f(x_1;\, w^{(2)}), & ~f(x_2;\, w^{(2)}), & ~\cdots, & ~f(x_N;\, w^{(2)}) \\
\vdots & \vdots & &  \vdots \\
f(x_1;\, w^{(M)}), & ~f(x_2;\, w^{(M)}), & ~\cdots, & ~f(x_N;\, w^{(M)})
\end{pmatrix}.
\end{equation*}
By the definition of N-shattering, every two rows in the above matrix are different. Thus for each $j\neq j'$, there exists some $i\in \{1,\dots,N\}$ such that $f(x_i;\, w^{(j)})\neq f(x_i;\, w^{(j')})$. By above arguments, there exists some binary function in $\mathcal{I}$ that takes different values on $(x_i,w^{(j)})$ and $(x_i, w^{(j')})$. Then there exists some $1\leq r\leq R$ such that the signs of $P_r(x_i, w^{(j)})$ and $P_r(x_i,w^{(j')})$ are different. In other words, each $j\in \{1,\dots,M\}$ gives a unique configuration of the signs of the $N\cdot R$ polynomials in $\mathcal{I}(x_1),\dots, \mathcal{I}(x_N)$. Utilizing Lemma~\ref{lem:nn_config} again, we obtain the desired result. 
\hfill $\square$

\bibliographystyle{apalike}
\bibliography{reference}

\end{document}